\documentclass[letterpaper, 10 pt, conference]{ieeeconf}  

\IEEEoverridecommandlockouts                              

\usepackage{hyperref}
\usepackage[table,xcdraw]{xcolor}

\definecolor{deep-red}{RGB}{192, 0, 0}

\usepackage{graphicx} 
\usepackage{comment}
\usepackage{color}
\usepackage{amsmath}
\usepackage{amsfonts}

\DeclareMathOperator*{\argmin}{arg\,min}

\usepackage{xcolor, soul}

\title{\LARGE \bf
Haptic-Assisted Collaborative Robot Framework for Improved Situational Awareness in Skull Base Surgery
}

\author{ Hisashi Ishida$^{1*}$, Manish Sahu$^{1*}$, Adnan Munawar$^1$, Nimesh Nagururu$^2$,  Deepa Galaiya$^{1, 2}$\\ Peter Kazanzides$^1$, Francis X. Creighton$^{1, 2}$, and Russell H. Taylor$^{1, 2}$ 
\thanks{$^1$Department of Computer Science, Johns Hopkins University, Baltimore, MD $21218$, USA. $^2$ Department of Otolaryngology-Head and Neck Surgery, Johns Hopkins University School of Medicine, Baltimore, MD $21287$, USA. *These authors contributed equally. Email: {\tt hishida3@jhu.edu}}%
}

\begin{document}

\maketitle
\thispagestyle{empty}
\pagestyle{empty}

\begin{abstract}
Skull base surgery is a  demanding field in which surgeons operate in and around the skull while avoiding critical anatomical structures including nerves and vasculature. While image-guided surgical navigation is the prevailing standard, limitation still exists requiring personalized planning and recognizing the irreplaceable role of a skilled surgeon.
This paper presents a collaboratively controlled robotic system tailored for assisted drilling in skull base surgery. Our central hypothesis posits that this collaborative system, enriched with haptic assistive modes to enforce virtual fixtures, holds the potential to significantly enhance surgical safety, streamline efficiency, and alleviate the physical demands on the surgeon. The paper describes the intricate system development work required to enable these virtual fixtures through haptic assistive modes.
To validate our system's performance and effectiveness, we conducted initial feasibility experiments involving a medical student and two experienced surgeons. The experiment focused on drilling around critical structures following cortical mastoidectomy, utilizing dental stone phantom and cadaveric models. Our experimental results demonstrate that our proposed haptic feedback mechanism enhances the safety of drilling around critical structures compared to systems lacking haptic assistance. With the aid of our system, surgeons were able to safely skeletonize the critical structures without breaching any critical structure even under obstructed view of the surgical site.
\end{abstract}

\section{INTRODUCTION}
Skull base surgery poses a profound challenge, demanding the utmost precision as surgeons dissect around critical structures, including nerves and vessels, which are often concealed by operable tissue at sub-millimeter distances~\cite{meybodi2023comprehensive}. Achieving this precision necessitates an intricate understanding of precise anatomy and sub-millimetric control during drilling—a skill acquired through extensive training and practice~\cite{gantz2018evolution}. Furthermore, surgeons must adapt their approach to accommodate the substantial anatomical variation observed among patients.
In light of this complexity, the integration of computer-aided and robotic assistance holds significant promise~\cite{zagzoog2018state}. Robotic assistance, in particular, offers the advantage of providing precise, tremor-free manipulation of surgical instruments. One pivotal component of many lateral skull base surgeries is mastoidectomy, or removal of mastoid temporal bone~\cite{bennett2006indications}, and robotic assistance has found a valuable role in this procedure ~\cite{danilchenko2011robotic}. However, mastoidectomy often represents only the initial step in these lateral skull base surgeries, and further drilling in the vicinity of critical structures is imperative to access the surgical targets including neoplasms and the organs of hearing. Consequently, robotic systems must evolve to encompass assistance throughout the entirety of skull base surgery~\cite{lim2016semi}.
Safe deployment of such robotic systems into the skull base requires the robot to develop situational awareness and safety-driven virtual fixtures (also known as active constraints)~\cite{bowyer2013active} to control its interaction with the surgical environment~\cite{simaan2015intelligent,attanasio2021autonomy}.

\begin{figure}[h]%
\centering
\includegraphics[width=.45\textwidth]
{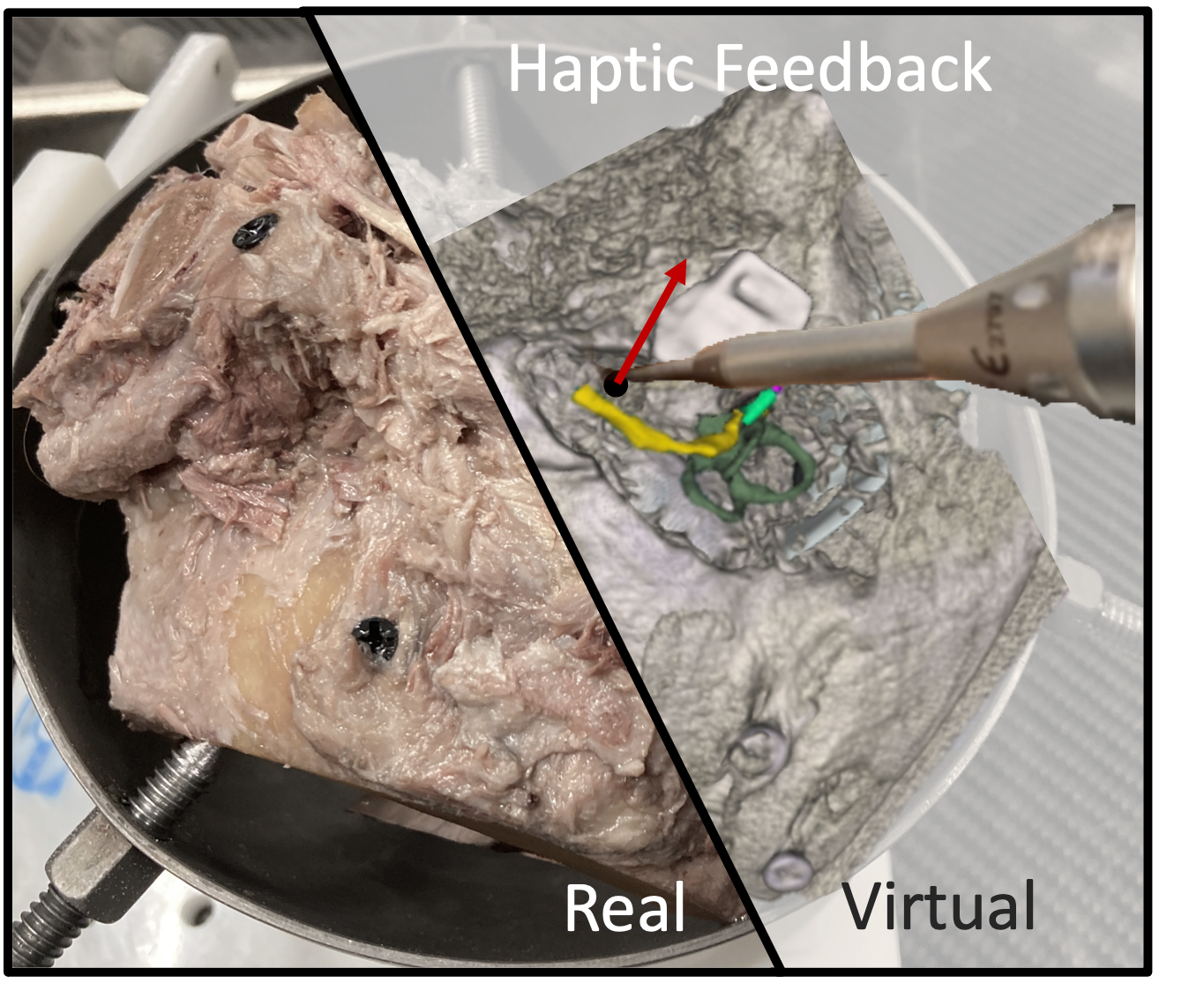}
\caption{
Conceptual overview of the safety-driven haptic assistance: Compliance force, defining the preferred direction of the drill based on the robot’s position relative to the anatomical surface, is computed in the dynamic simulation model. This force is fed back to the collaborative robot controller to generate haptic feedback. 
}\label{fig:Main_Figure}
\vspace{-5pt}
\end{figure}

The primary objective of this work is to enhance surgical situational awareness by augmenting human perception through the provision of safety-driven haptic assistance during the removal of soft tissues surrounding critical structures in the post-mastoidectomy phase of robotic-assisted surgery (Fig. \ref{fig:Main_Figure}).
A major challenge in enabling such haptic assistance lies in generating appropriate feedback proportional to the hand force applied by the surgeon to guide the robot while computationally compensating for the gravitational forces exerted by the drill mounted on the robot. This work outlines the system development process and pipeline required to facilitate haptic feedback during skull base drilling using a cooperative robot. This encompasses the seamless integration of the cooperative robot with a real-time simulation-driven navigation system, as well as the computational elimination of gravitational forces attributed to the drill's weight. Moreover, the generation of guidance virtual fixtures is derived directly from patient volumetric imaging data to provide patient-specific assistance. 

To demonstrate the performance and efficacy of our system, we conducted initial feasibility experiments involving a medical student and two experienced surgeons. 
Surgeons drilled around critical structures following cortical mastoidectomy in phantom and cadaveric models.
Our experimental results demonstrate that the proposed feedback mechanism enhances the safety of drilling around critical structures when compared to a system lacking haptic assistance.

The key contributions of the work are:
\begin{enumerate}
    \item A comprehensive pipeline that spans from creation of complex anatomical constraints to intraoperative collaborative robotic assistance.
    \item The development and seamless integration of safety-driven haptic assistance to facilitate skull base drilling.
    \item Initial feasibility experiments to showcase the system's accuracy and performance, particularly in scenarios involving intricate anatomical features 
    such as the facial nerve.
\end{enumerate}
    
It is important to note that although our system is specifically designed for collaborative robotic assistance in skull base surgery, it is constructed using open-source libraries and adheres to industry-standard data formats. Consequently, our approach has the potential to be applicable across a wide range of robotic systems, surgical procedures, and anatomical regions.

\section{Related Work}

Multiple works have introduced robotic systems designed to assist humans in various manipulation tasks \cite{bowyer2013active}. 
These systems can be broadly categorized based on the degree of automation they offer, ranging from fully automated to semi-autonomous control. For the sake of brevity, we specifically consider robotic systems applicable to head and neck procedures.
While autonomous bone drilling systems have shown promise~\cite{dillon2016making}, it necessitates precise planning for every possible scenario and surgical approach. It lacks the invaluable input of a surgeon's procedural knowledge, technical skill, and real-time perceptual feedback. In contrast, cooperative systems involve both the surgeon and the robot simultaneously holding the drill. This arrangement allows the drill to move in proportion to the surgeon's applied force while adhering to active constraints set by the controller. Safety constraints can be activated when the drill nears a critical structure and remain inactive otherwise. In our work, we employ a cooperatively controlled robot for robotic-assisted drilling.

Within the context of safety-driven robotic assistance, Ding et al.~\cite{ding2021volumetric} and Xia et al.~\cite{Xia2008} implemented planar virtual constraints in their control system and demonstrated its utility in phantom drilling. However, planar constraints, while useful for defining the robot's workspace, cannot model anatomical structures which are inherently non-planar.
Other studies have used segmented CT scans to derive complex anatomical constraints directly, enforcing a safety margin of 3 mm between critical structures and the drill ~\cite{lim2016semi,yoo2017cadaver}. While these works showcased the utility of enforcing safety margins with a cooperative robot during mastoidectomy procedures, these safety margins were frequently breached, resulting in additional time spent reactivating the robot and suboptimal user experiences.

Our paper focuses on providing haptic feedback to effectively guide the drill motion away from critical structures, instead of enforcing hard safety barriers. Unlike forbidden-region-based virtual fixtures, which aim to keep the manipulator's tool away from restricted areas, our haptic-based virtual fixtures assist the user in moving the robot manipulator along desired paths or surfaces\cite{abbott2007haptic,bowyer2013active}. To achieve this, we derive distance fields~\cite{ishida2023improving} from patient volumetric imaging data. These distance fields define the preferred direction of the drill based on the robot's position relative to the desired surface. By restraining user-commanded motions in non-desired directions, we establish passive guidance, effectively steering the drill away from critical structures.

\section{Methodology}
\begin{figure*}[]
\centering
\includegraphics[width=0.8\textwidth]{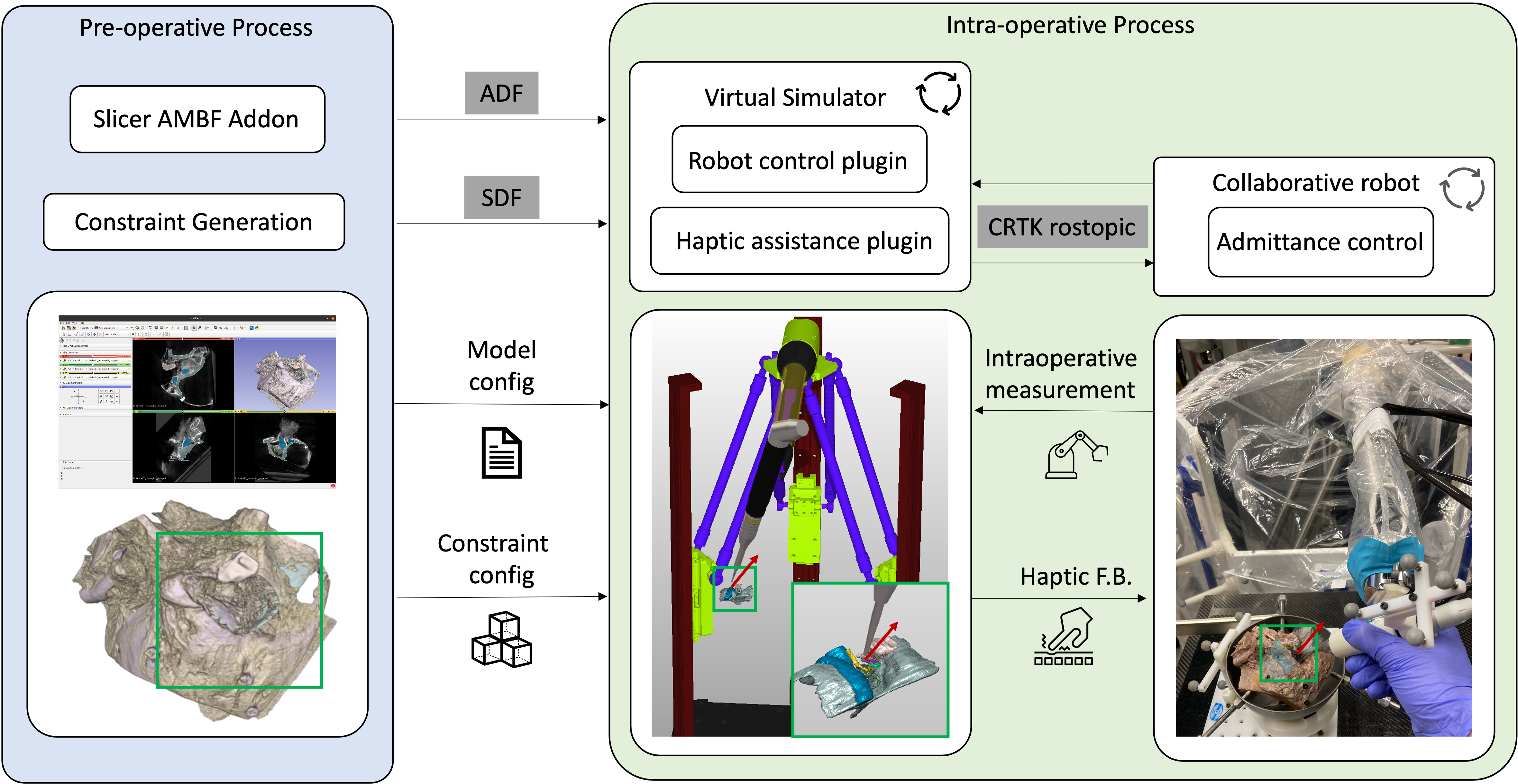}
\caption{System Overview. (Left) preoperative creation of patient model and constraints configuration. (Middle) intra-operative dynamic simulation of the robot and anatomy as well as generation of compliance force. (Right) Collaborative robot receiving the compliance force and augment the safety of the procedure.  }\label{fig:System_Overview}
\end{figure*}



At the heart of our system, we have a cooperatively controlled robot and an interactive dynamic simulation environment, Asynchronous Multi-Body Framework (AMBF)~\cite{AMBF}. 
This simulation environment comprises three primary components: a patient anatomical model, a collaborative robot, and a surgical drill. The anatomical model is crafted from segmented preoperative CT images and then registered with the actual patient anatomy and the robot's workspace. This ensures that the virtual representation seamlessly mirrors the real-world anatomical structures and its safe operational boundaries. During operation, the simulation model continuously gathers data through the robot's state and optical tracking. Simultaneously, it offers real-time situational awareness by establishing spatial relationships between surgical instruments and the surrounding tissues. This dynamic model forms the bedrock for implementing virtual fixtures that envelop critical anatomical structures, effectively guiding the surgeon's actions while ensuring precision and safety during the procedure.


In Section \ref{sec: system_overview}, the overview of the system (Fig. \ref{fig:System_Overview}) and interaction between its components are described. The constraints are modeled using the Signed Distance Field (SDF) for the anatomical volumes (Section \ref{sec: sdf_calculation}). The ``Haptic assistance plugin'' monitors the distance between the virtual instruments and different anatomies in real time and generates the compliance force based on SDF (Section \ref{sec: haptic_feedback}). The compliance force is fed into the collaboratively controlled robot to produce a sensation of haptic feedback (Section \ref{sec: REMS}). Lastly, we discuss the calibration and registration pipeline that we used.

\subsection{System overview}
\label{sec: system_overview}
From the preoperative CT scans, our system generates SDF for critical anatomies and provides real-time haptic feedback. This incorporates intricate anatomical structures, mitigating collision risks with anatomies and enhancing the safety of the operator.
The process involves the use of 3D Slicer \cite{FEDOROV20121323}, a widely adopted imaging software, for loading and annotating CT scans. The AMBF Slicer plugin is employed to convert the ``seg.nrrd'' format into AMBF Description Files (ADF) compatible with the virtual simulator.  
Once the SDF for all segmented anatomies is generated, the ``Haptic assistance plugin'' calculates the proposed haptic feedback and sends it back as a compliance force to the robot.


Effective communication between the real environment and the simulator is realized by adopting the Collaborative Robotics Toolkit (CRTK) \cite{Su2020} convention, promoting modularity and seamless integration with other robotic systems.


We use the robot's built-in admittance controller while only incorporating the SDF-based force feedback from the simulator for better interoperability with our system.

\subsection{SDF calculation and volume representation}
\label{sec: sdf_calculation}
An SDF volume for a specific anatomical structure, $\mathbb{a}$, is represented as a 3D voxel grid, $\mathbb{S}_\mathbb{a}$, where each voxel stores the distance between the voxel and the closest voxel containing this specific anatomic structure. In this grid, positive values indicate voxels outside the anatomy, while negative values indicate voxels on the inside. In this paper, we adopts the method proposed in our previous work \cite{ishida2023improving}, which internally facilitates Saito and Toriwaki's method \cite{saito_new_1994} to calculate SDF in a parallelized manner.

From the SDF volume, $\mathbb{S}_\mathbb{a}$, the closest distance between the drill tip, $x \in \mathbb{R}^3$, and the nearest anatomy is calculated,$d_\mathbb{a} (x) \in \mathbb{R}$, and using the finite difference, the direction from the drill tip to the closest point of the nearest anatomy, $\Vec{d}_\mathbb{a} (|\Vec{d}_\mathbb{a}| = 1)$, is derived. Please refer to \cite{ishida2023improving} for further details.

\subsection{SDF-based haptic feedback}
\label{sec: haptic_feedback}
Given $n \in \mathbb{Z}$ critical anatomies in the surgical scene, the SDF-based force feedback, $\Vec{F}_{SDF}\in \mathbb{R}^3$, can be written as:
\begin{equation}
\vec{F}_{SDF}^{(n)} =
    \begin{cases}
        F_{max} \Vec{d}_\mathbb{a} & \text{if } d_\mathbb{a} < \tau_{0}^{(n)}\\
        F_{max} \exp{\lambda \left(\tau_0^{(n)} - d_\mathbb{a} \right)}\Vec{d}_\mathbb{a} & \text{if } \tau_{0}^{(n)} < d_\mathbb{a} < \tau_{f}^{(n)}\\
        \Vec{0} & \text{if } \tau_{f}^{(n)} < d_\mathbb{a} 
    \end{cases}
\end{equation}
\begin{equation}
    \Vec{F}_{SDF} = \sum_n \vec{F}_{SDF}^{(n)}
\end{equation}
where $\vec{F}_{SDF}^{(n)} \in \mathbb{R}^3$ represents the force from the $n$th anatomy in the scene. The closest distance to the $n$th anatomy, $d_\mathbb{a}$, and the direction toward the nearest point in the $n$th anatomy, $|\Vec{d}_\mathbb{a}|$, is derived from the previous section. $\tau_0^{(n)} \in \mathbb{R}$ and $\tau_{f}^{(n)} \in \mathbb{R}$ are the thresholds to activate the hard constraint and haptic feedback, respectively. $F_{max}$ is the maximum force in newtons, [N], and $\lambda \in \mathbb{R}$ is a decay constant that determines how steeply the force increases when the virtual drill is close to the anatomy.

\begin{figure}[t]%
\centering
\includegraphics[width=0.47\textwidth]{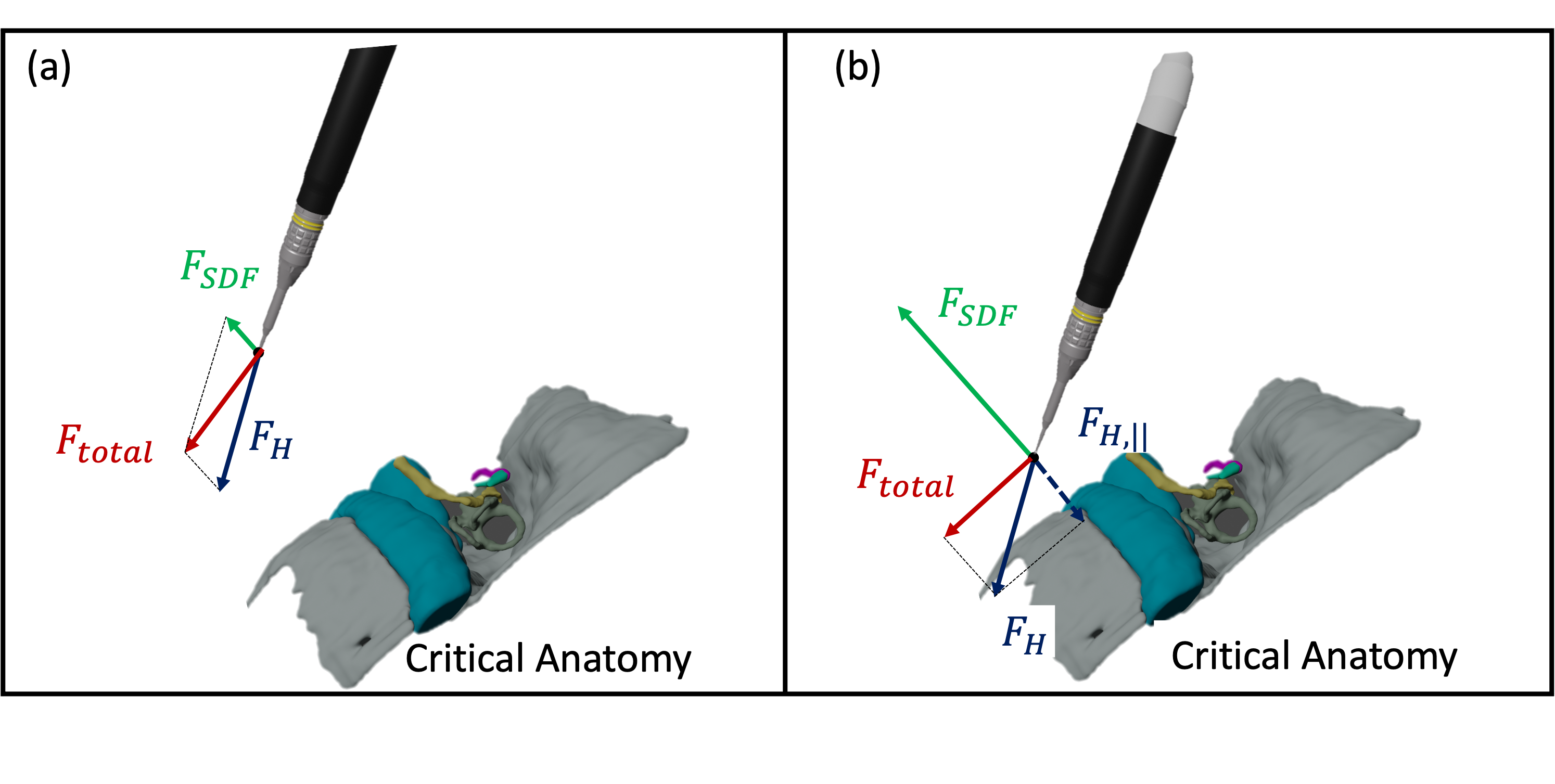}
\setlength{\abovecaptionskip}{-10pt}
\caption{
Here, $F_H$ denotes the user-applied force and $F_C$ is the compliance force generated using SDF. The compliance force is proportional to the user-applied force and effectively guides the motion in the preferred direction.}

\label{fig:SDFbasedVF}
\end{figure}

To maintain the usability of the cooperative robots, we adjust the SDF-based force feedback, $\Vec{F}_{SDF}$, according to the user-applied force, $\Vec{F}_{H} \in \mathbb{R}^3$ by setting $F_{max} = |\Vec{F}_{H}|$.
Additionally, the compliance force, $\Vec{F}_{C} \in \mathbb{R}^3$, that will be sent to the collaborative robot is adjusted using the following rule:
\begin{equation}
    \begin{cases}
        \Vec{F}_{C} = \Vec{F}_{SDF} & \text{if } (\Vec{F}_{H} + \Vec{F}_{SDF}) \cdot  \Vec{F}_{H,||} < 0\\
        \Vec{F}_{C} = -\Vec{F}_{H,||} & \text{Otherwise}
    \end{cases}
    \label{eq: ForceLimit}
\end{equation}

where $\Vec{F}_{H,||} \in \mathbb{R}^3$ is the component of the hand force that is parallel to $\Vec{F}_{SDF}$. Equation (\ref{eq: ForceLimit}) regulates the SDF-based force so that the total force ($F_H + F_C$) that controls the robot remains aligned with the direction of the user's applied force (Fig. \ref{fig:SDFbasedVF}). 

\subsection{Virtual simulator and robotic system.}
\subsubsection{Virtual simulation}
AMBF successfully demonstrated the virtual reality simulation for skull base surgery~\cite{FIVRS2023,shu2023twin}. We further develop our system to generate compliance force that is later utilized in the admittance control for the collaborative robot to enable haptic feedback. First, we created the virtual robot that accurately represents the real robotic system. Next, we import the virtual robot, anatomy from the CT, and SDF for all critical anatomies in this simulator. The SDF-based active constraint explained in the previous section is enforced using the plugin functionality (SDF assistance plugin), and the virtual robot motion is synchronized by the robot control plugin.

\subsubsection{Collaborative robotic system}
\label{sec: REMS}
The Robotic ENT (Ear, Nose, and Throat) Microsurgery System (REMS) is a cooperatively-controlled robot created specifically for use within otolaryngology–head and neck surgery.~\cite{olds2014preliminary,olds2015phd} For this study, we use a pre-clinical version developed by Galen Robotics (Galen Robotics, Baltimore, MD). REMS offers a significant benefit in instrument stability in head and neck surgery. The cooperative mode of the robot uses the admittance control law:
\begin{equation}
    \argmin_{\Delta q} \left( |GF_{H} - J\Delta q|\right)
\end{equation}
$G\in \mathbb{R}^{6 \times 6}$ is a diagonal matrix that represents the admittance gains, $J \in \mathbb{R}^{6 \times m}$ is a Jacobian and $\Delta q \in \mathbb{R}^m$ is a joint velocity vector. Thus, the incremental motion of the robot end-effector, $\Delta x \in \mathbb{R}^6$, is expressed as ($\Delta x = J \Delta q$). 

We introduced a compliance force term, denoted as $F_C$, to the admittance control for the robot's motion. As a result, the updated optimization equation can be expressed as:

\begin{equation}
    \argmin_{\Delta q} \left( |G(F_H + F_C) - J\Delta q|\right)
\end{equation}

\subsection{Calibration and registration}

Our system requires accurate spatial representation between the physical system components for which we employ state-of-the-art registration and calibration algorithms.

For better transparency of the collaborative robot, it is critical to filter only the force from the user. This necessitates the exclusion of external forces, a primary example being the gravitational impact on the drill. Addressing this concern is challenging problem due to the flexible cable of the drill, which exhibits non-linearity. We employed a model for this external force based on a Bernstein polynomial and incorporated gravity compensation as described in \cite{chen2023force}.

\begin{figure}[ht]%
\centering
\includegraphics[width=0.45\textwidth]{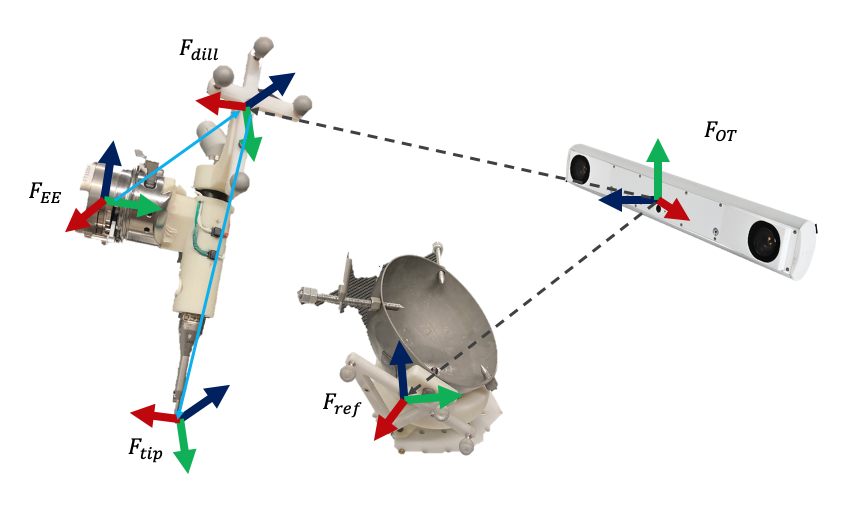}
\setlength{\abovecaptionskip}{-10pt}
\caption{Frame transform diagram. $F_{OT}$ represents the frame for the optical tracker. $F_{drill}$, $F_{ref}$ are the frames attached to the drill marker and reference marker, respectively. $F_{EE}$ is the frame for the robot end-effector.}\label{fig:Frame_Transformation}
\end{figure}

The registration between the actual environment and the simulation is fundamental in this work, especially in locating the drill tip with respect to the robot . To achieve sub-millimeter precision, we utilize the optical tracker system, FusionTrack 500 (Atracsys, Switzerland) (Fig. \ref{fig:Frame_Transformation}). It offers a mean tracking error of $0.02$ ($\pm 0.02$) $mm$.~\cite{shu2023twin}

We perform a hand-eye calibration to calculate the transformation between $F_{EE}$ and $F_{drill}$. This hand-eye calibration is followed by a pivot calibration to get the position of the drill tip, $F_{tip}$, with respect to the marker attached to the drill ($F_{drill}$).
To render haptic feedback that aligns with real-time anatomical positioning, the anatomy is registered with respect to the robot using point-set registration technique.


\section{Evaluation and results}
To demonstrate the applicability of our system and assess its efficacy, two experiments were performed: one utilizing a phantom with dental stone and the other involving a cadaver temporal bone.
The dental stone experiment primarily served to validate our system in a controlled setting and also provided an acclimation period before the temporal bone experiment.

\begin{table}[h]
\begin{center}
\caption{Haptic related control Parameters}\label{table:param}
\scalebox{1.1}{
\begin{tabular}{|c|c|c|c|c|c|}
\hline
\multicolumn{3}{|c|}{Denatal stone experiment} &
\multicolumn{3}{c|}{Temporal bone experiment} \\

\hline 
$\tau_0 ~[mm]$ & $\tau_f ~[mm]$ & $\lambda$ & $\tau_0 ~[mm]$ & $\tau_f ~[mm]$ & $\lambda$ \\
\hline
1.0 & 4.0 & 1.0 & 0.5 & 4.0 & 2.0\\
\hline
\end{tabular}}
\end{center}
$\tau_0 $: distance where the unpreferred direction is fully constrained\\
$\tau_f$: distance where proposed haptic feedback activates\\
$\lambda$: decay constant for the proposed haptic feedback
\end{table}

The relevant parameters for the proposed haptic feedback are described in Table \ref{table:param}. 
During the dental stone experiment, we opted for a $\tau_0 = 1 mm$, which was subsequently reduced to$\tau_0 = 0.5 mm$ in the temporal bone experiments. Furthermore, $\lambda$ is also tuned for the cadaveric experiment to ensure that the surgeons can approach closer to the critical anatomy.

\subsection{Experiment setup}
\begin{figure}[ht]%
\centering
\includegraphics[width=.45\textwidth]{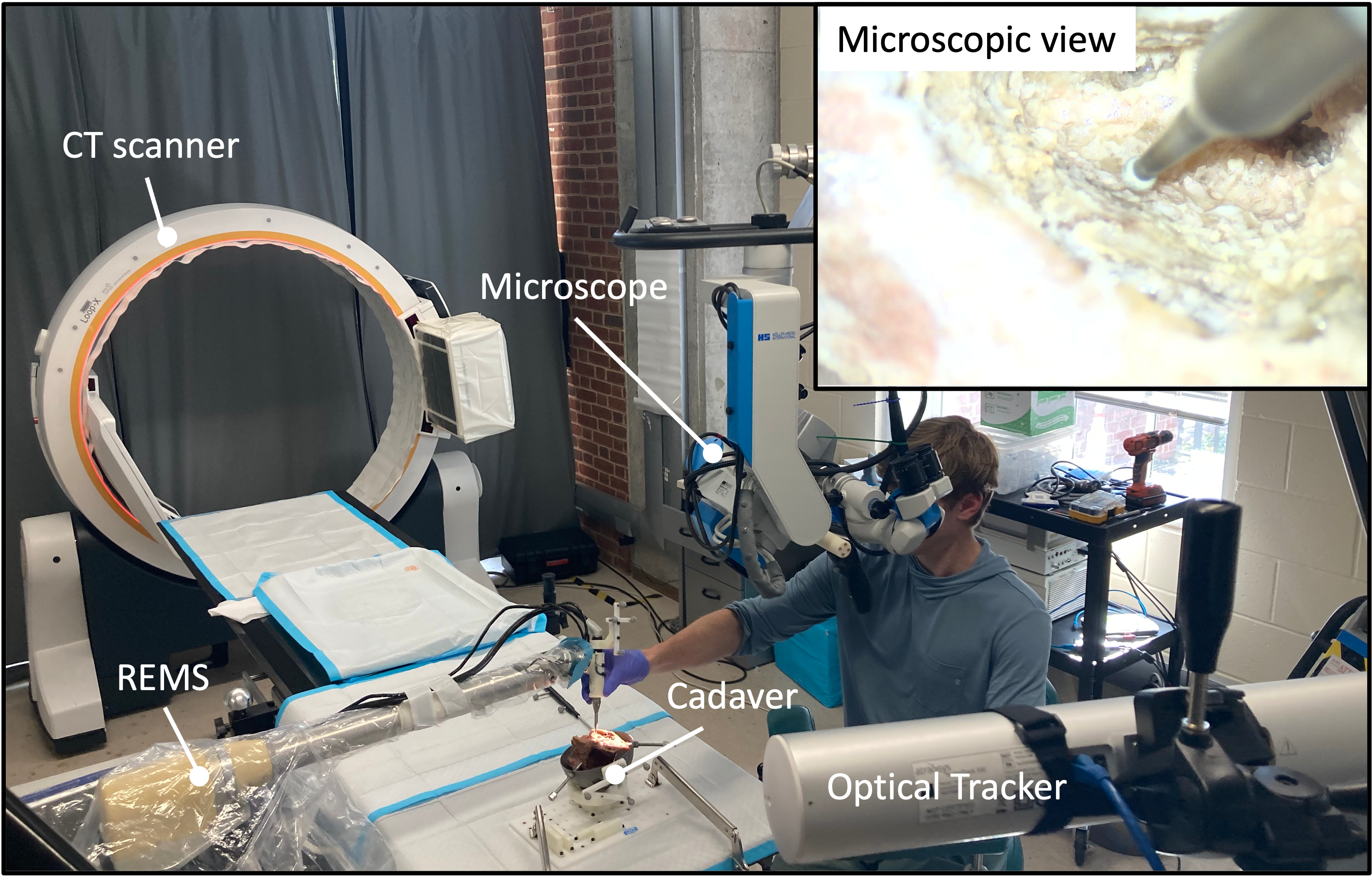}
\caption{Experimental setup. Surgeon uses the surgical drill attached to the robot under microscopic view. Optical tracker was located next to the robot to monitor both the drill and the anatomy.}\label{fig:Experimental_Setup}
\end{figure}

To prepare for the experiment, we first affixed registration pins to the cadaveric temporal bones. High-resolution CBCT (Brainlab LoopX, $0.26\,mm^3$ voxel size) scans were then taken for each temporal bone. Following this, we used 3D Slicer to annotate critical structures and the locations of the registration pins.
These annotated anatomical structures were later used to create the patient anatomical model and constraints for our system.
Prior to the experiment, the temporal bone was securely positioned within a temporal bone holder. The holder was then firmly anchored to a surgical table to prevent any potential inadvarent movement. Tracking markers were affixed to both the surgical drill and the temporal bone holder to monitor any movements or registration discrepancies. Next, we initiated the calibration and registration process. The hand-eye calibration process resulted in an RMSE of approximately 0.2 $mm$ for translation and 0.3 degrees for rotation. For pivot calibration, we used a 2 $mm$ drill tip, achieving an RMSE value of 0.03 $mm$.
Finally, the temporal bone was registered to the virtual simulation using a point-set registration method. These points were carefully sampled using the surgical drill, leading to an RMSE of less than 0.5 $mm$.
During the experiment, surgeons conducted drilling procedures around critical structures under a surgical microscope (Haag-Streit, Köniz, Switzerland) equipped with stereo vision. Following the experiment, CT scans were taken for postoperative evaluation (Fig. \ref{fig:Experimental_Setup}).

\subsection{Dental stone experiment}
\subsubsection{Experiment design}

\begin{figure}[t]%
\centering
\includegraphics[width=0.47\textwidth]{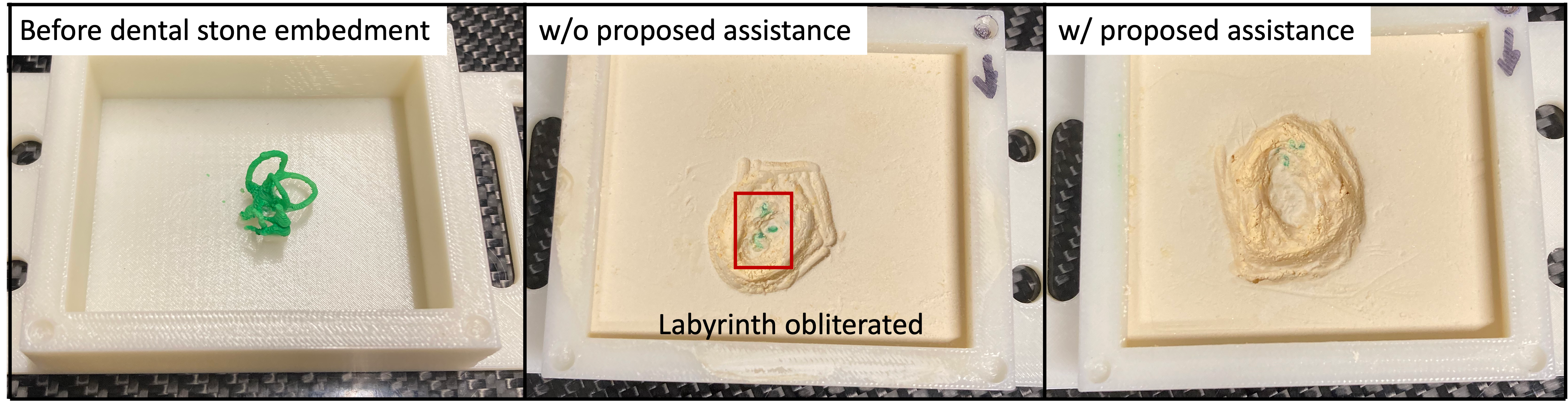}
\caption{Example visualization of dental stone experiment with a medical student. (Left) 3D printed labyrinth, (middle) Result without assistance. Part of labyrinth was inadvertently drilled. (Right) Result with assistance. No injury to the labyrinth was observed. }\label{fig:DSReuslt}
\end{figure}

To validate the safety-enhancing capabilities of our proposed method, we designed a phantom experiment using dental stone powder. A 3D printed bony labyrinth, an inner ear structure,  was embedded within the dental stone phantom (as depicted in Fig. \ref{fig:DSReuslt}). To ensure clear differentiation, the labyrinthine structure was painted in green, with the paint mixed with CT opaque material to facilitate high-quality segmentation.


A medical student and two attending surgeons were tasked with delicately skeletonizing the superior part of the labyrinth, aiming to avoid any damage. They were provided a time limit of 5 minutes to complete this task, performing it both with and without the assistance of our proposed method. We randomized the order to mitigate the learning effect. CT scans were taken before and after the experiment, and any damage incurred to the labyrinth was assessed.

\subsubsection{Result and discussion}



\begin{table}[h]
\begin{center}
\caption{Results from dental stone experiment}\label{table:DS_result}
\begin{tabular}{|c|c|c|c|c|}
\hline
& \multicolumn{2}{c|}{Damage on anatomy~[$mm^3$]}& \multicolumn{2}{c|}{Drilled volume~[$mm^3$]}\\
\hline
subject & w/o VF & w/ VF & w/o VF & w/ VF \\
\hline
S1 & 0.58 & 0.0 & 878.4 & 1406.4 \\  
A1 & 0.0 & 0.0 & 694.7 & 726.6\\ 
A2 & 0.0  & 0.0 & 511.3 & 536.8 \\
\hline
\end{tabular}
\end{center}
\end{table}

Table \ref{table:DS_result} shows the quantitative result for the dental stone experiment. 
The results show that the medical student inadvertently breached a critical structure when operating without haptic feedback, whereas no breaches occurred with haptic assistance (Fig. \ref{fig:DSReuslt}). In contrast, experienced surgeons were able to avoid the critical structure regardless of the presence of the haptic feedback.
Further analysis on the drilled volume suggests that experienced surgeons approach the critical structure more carefully than medical trainees. The drilled volume remained in a similar range for both surgeons with and without haptic assistance. Nevertheless, the medical student greatly benefited from the proposed haptic assistance, presumably becoming more confident in safe drilling. This is evident as the volume drilled increased by approximately 1.6 times compared to when no haptic assistance was enforced.

\setcounter{figure}{7} 
\begin{figure*}[h]%
\centering
\includegraphics[width=0.95\textwidth]{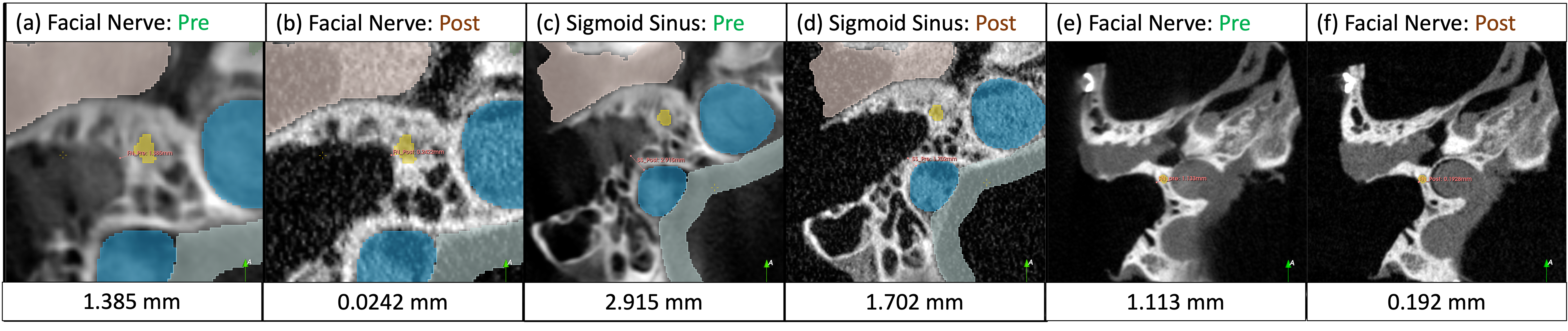}
\setlength{\abovecaptionskip}{-5pt}
\caption{Comparison between pre- and post-operative CT scans for closest anatomies. First surgeon (a-d). Second surgeon (e-f). The second surgeon only approached the facial nerve. Both surgeons were able to skeletonize the facial nerve and sigmoid sinus without damaging them. The red lines in the figure represent the closest distance between the critical structure and the exposed surface. }\label{fig:A1ResultCTs}
\end{figure*}

\setcounter{figure}{6} 
\begin{figure}[htp]%
\centering
\includegraphics[width=0.47\textwidth]{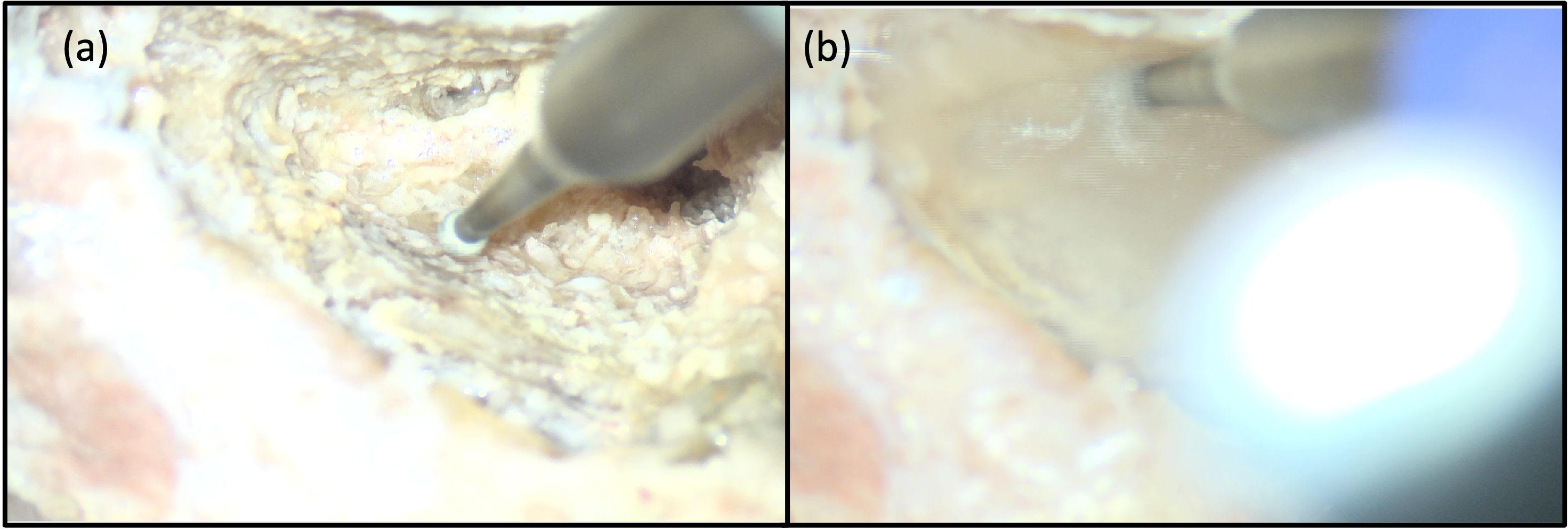}
\setlength{\abovecaptionskip}{-2pt}
\caption{Example microscope view. (a) Normal view, (b) with occulusion.}
\label{fig:Occuluded View}
\end{figure}

\subsection{Temporal bone  experiment}
\subsubsection{Experimental design}

To further assess the efficacy of haptic assistance, we conducted experiments on cadaveric temporal bones with the two experienced surgeons. This cadaveric study was designed to closely replicate real-life surgical scenarios, thereby facilitating a more accurate evaluation of the proposed method. In this experiment, the surgeons were tasked with the challenging process of skeletonizing critical structures under conditions that severely limited their visual cues. This was achieved by introducing water into the surgical site during the drilling procedure (as depicted in Fig. \ref{fig:Occuluded View}). This deliberate reduction in visibility within the complex temporal bone anatomy presented challenges akin to the irrigation process encountered in actual surgical scenarios.
It is worth noting that the temporal bone had previously undergone mastoidectomy, emphasizing that the drilling primarily occurred within the proximity of critical structures.


\subsubsection{Results and discussion}
As demonstrated in our earlier dental stone experiments, experienced surgeons exhibited a high level of precision and were adept at avoiding damage to critical structures under normal conditions in cadaveric experiment. However, when confronted with such limited visibility, characteristic of the intricate temporal bone anatomy, their primary reliance shifted to haptic feedback while navigating around critical structures.
Despite in those challenging conditions, both surgeons successfully navigated and drilled around the intricate anatomical structures without causing any damage, thereby highlighting the practicality and effectiveness of our haptic assistance system. As depicted in Fig. \ref{fig:A1ResultCTs}, the results indicate that the surgeons successfully skeletonized the facial nerve and the sigmoid sinus without any damage.
This result shows that there is significant potential for this system to be effectively implemented in actual surgical procedures. Such potential not only highlights the system's technical capabilities but also highlights its relevance in advancing surgical precision and safety using the collaborative robot.




\section{Summary and Future Work}

In this work, we have developed a collaborative robot system to enhance situational awareness in skull base surgery by enhancing human perception through safety-driven haptic assistance. Our system ensure safe drill manipulation by providing progressively increasing haptic feedback as the drill approaches critical structures. Initial experiments using dental stone phantoms and cadaveric temporal bones demonstrate the system's feasibility and effectiveness for both novice and experienced surgeons.

While our work represents a significant advancement, certain limitations warrant attention. First, our current pipeline relied on manual segmentation for generating patient anatomical models and anatomical constraints. Future work involves integrating an automated segmentation pipeline~\cite{ding2023self} to streamline this process.
Furthermore, our system relies on high-resolution radiological scans to create precise anatomical models and constraints. In cases where high-resolution patient scans are unavailable, recent methods like those presented in~\cite{zhang2023Implicit} can be considered as an alternative for SDF.
Secondly, our experiments employed invasive fiducial markers for registration, ensuring a high level of accuracy. Ongoing efforts concentrate on integrating vision-based tracking and registration algorithms to minimize invasiveness and enhance precision.
Lastly, despite our promising initial experimental results, the system's performance necessitates validation through extensive user studies conducted on a larger scale. This remains a focus of our future research endeavors.

In summary, our work contributes to the advancement of robotic integration in skull base procedures, paving the way for enhancing surgical precision while preserving the indispensable role and expertise of the surgeon.

\section*{Acknowledgments and Disclosures}
Hisashi Ishida was supported in part by the ITO foundation for international education exchange, Japan. Nimesh Nagururu is supported in part by NCATS TL1 Grant TR003100.
This work was also supported in part by a research contract from Galen Robotics, by NIDCD K08 Grant DC019708, by a research agreement with the Hong Kong Multi-Scale Medical Robotics Centre,
and by Johns Hopkins University internal funds.
Under a license agreement between Galen Robotics, Inc and the Johns Hopkins University, Russell H. Taylor and Johns Hopkins University are entitled to royalty distributions on technology that may possibly be related to that discussed in this publication. Dr. Taylor also is a paid consultant to and owns equity in Galen Robotics, Inc. This arrangement has been reviewed and approved by Johns Hopkins University in accordance with its conflict-of-interest policies.

\section*{Supplementary information} 
A supplementary video is provided with the submission. For more information, visit the \href{https://github.com/LCSR-CIIS/sdf_virtual_fixture/tree/doubleloop}{project repository}.


\bibliography{bibliography}
\bibliographystyle{IEEEtran}

\end{document}